\documentclass[copyright,creativecommons]{eptcs}
\providecommand{\event}{ICLP 2019} 
\usepackage{underscore}
\usepackage{latexsym}
\usepackage{enumitem}
\usepackage[pdftex]{graphicx}
\usepackage{wrapfig}
\usepackage[noend]{algorithmic}
\usepackage{listings}
\usepackage[x11names,svgnames]{xcolor}
\usepackage{adjustbox}

\graphicspath{{images/}}

\newcommand{\comment}[1]{}

\newtheorem{definition}{Definition} 
\newtheorem{example}{Example} 

  \title{Value of Information in Probabilistic Logic Programs\thanks{This work was supported in part by NSF Grant IIS-1447549.}}

  \author{Sarthak Ghosh  \qquad\qquad C. R. Ramakrishnan
 \institute{Department of Computer Science\\
  Stony Brook University\\
  Stony Brook, NY 11794, USA}
  \email{saghosh@cs.stonybrook.edu \qquad\qquad cram@cs.stonybrook.edu}
    }   

\def\titlerunning{VoI in PLPs}
\def\authorrunning{S.\ Ghosh \& C.\ R.\ Ramakrishnan}

\begin{document}


\maketitle

  \begin{abstract}
    In medical decision making, we have to choose among several expensive diagnostic tests such that the certainty about a patient’s health is maximized while remaining within the bounds of resources like time and money. The expected increase in certainty in the patient's condition due to performing a test is called the value of information (VoI) for that test. In general, VoI relates to acquiring additional information to improve decision-making based on probabilistic reasoning in an uncertain system. This paper presents a framework for acquiring information based on VoI in uncertain systems modeled as Probabilistic Logic Programs (PLPs). Optimal decision-making in uncertain systems modeled as PLPs have already been studied before. But, acquiring additional information to further improve the results of making the optimal decision has remained open in this context.

    We model decision-making in an uncertain system with a PLP and a set of top-level queries, with a set of utility measures over the distributions of these queries.  The PLP is annotated with a set of atoms labeled as ``observable''; in the medical diagnosis example, the observable atoms will be results of diagnostic tests.  Each observable atom has an associated cost.  This setting of optimally selecting observations based on VoI is more general than that considered by any prior work.  Given a limited budget, optimally choosing observable atoms based on VoI is intractable in general.  We give a greedy algorithm for constructing a ``conditional plan'' of observations: a schedule where the selection of what atom to observe next depends on earlier observations.  We show that,  preempting the algorithm anytime before completion provides a usable result, the result improves over time, and, in the absence of a well-defined budget, converges to the optimal solution.
  \end{abstract}


\section{Introduction}
\paragraph{\textbf{Background.}} Probabilistic Logic Programs (PLPs) have been proposed as an expressive mechanism to model and reason about systems combining logical and statistical knowledge. 
Programming languages and systems
studied under the framework of
PLP include PRISM~\cite{sato:ijcai}, Problog~\cite{deraedt:ijcai},
PITA~\cite{riguzzi:tplp} and Problog2~\cite{dries:ecai}.  These
languages have similar declarative semantics based on the
\emph{distribution semantics}~\cite{sato:jair}.  At a high level, programs in these languages specify independent probabilistic choices among facts or rule instances, analogous to those in ICL~\cite{PooleICL}.  The joint distribution among these choices yields a distribution over non-probabilistic logic programs (also known as \emph{worlds} in the literature).  A distribution over definite logic programs induces a distribution over their least models; the probability of a query answer is the probability of the set of all least models that have the answer.

\paragraph{\textbf{The Driving Problem.}} In this paper, we consider the problem of \emph{data acquisition} in systems specified as  PLPs.  For instance, a PLP encoding of a medical diagnostic system may specify rules for potential diagnoses based on symptoms and test results, with distributions specified over the
rules, as well as symptom and test results.  Consider a case where we know only the symptoms presented by a patient, and want to know what tests to perform that will increase our confidence in the ultimate diagnosis.  These tests may be expensive, and a natural question here is \emph{which} tests yield data that \emph{best} influence the decisions, while fitting within an overall budget.   The expected increase in the quality of decisions enabled by the new data due to a test is called the value of information (VoI) for that test.  

\paragraph{\textbf{Technical Approach.}}
The VoI optimization problem has been studied in the past in restricted contexts of influence diagrams and probabilistic graphical models.  In this paper, we pose this optimization problem in the more expressive context of PLPs.  Without loss of generality, our technical development uses ProbLog as the underlying specification language for the combined logical/statistical system.  We use a simple sensor placement problem, described below, and the corresponding ProbLog program, in Fig.~\ref{fig:intro-example}, as a running example to illustrate our technical approach.

\begin{example}[\textnormal{\textbf{Monitoring the Temperature}}]\label{ex:voi} \rm
    There is a building with three rooms whose respective indoor temperatures are modeled by three random variables, $T_{1}$, $T_{2}$, and $T_{3}$; each $T_i$ is modeled by predicates \texttt{room($i$, $T$)} in the ProbLog program in Fig.~\ref{fig:intro-example}.  The reading of each sensor can either be `\texttt{hi}', indicating high temperature, or `\texttt{lo}', indicating low temperature. The sensors are used to adjust the heating inside the building such that the indoor temperature never drops too low. If at any point of time, the reading of any one sensor is `\texttt{lo}', the heating needs to be increased (predicate \texttt{heater\_on} in 
 Fig.~\ref{fig:intro-example}).
    
    Let the random variables $T_{1}$, $T_{2}$, and $T_{3}$ form a Bayesian network where $T_{2}$ is the child of $T_{1}$, and $T_{3}$ is the child of $T_{2}$. Let the probability distributions associated with this Bayesian network be as follows:
    \begin{quote}
        $\textit{Pr}(T_{1} = \textnormal{lo}) = 0.5$
        
        $\textit{Pr}(T_{2} = \textnormal{lo} | T_{1} = \textnormal{lo}) = \textit{Pr}(T_{3} = \textnormal{lo} | T_{2} = \textnormal{lo}) = 0.7$
        
        $\textit{Pr}(T_{2} = \textnormal{lo} | T_{1} = \textnormal{hi}) = \textit{Pr}(T_{3} = \textnormal{lo} | T_{2} = \textnormal{hi}) = 0.3$
    \end{quote}
These dependencies and distributions are encoded in the ProbLog program by the (probabilistic) facts defining \texttt{room($i$, $T$)}.  In this program, query \texttt{probability(heat\_on)} ($=0.755$) gives the likelihood that at least one room has low temperature, but \emph{under the condition that we have not measured any room's temperature.}

\begin{figure}[h]
\centering
\begin{minipage}{0.65\columnwidth}
\begin{center}
\framebox{
    \begin{minipage}{\columnwidth}
    \ttfamily\small
    \begin{algorithmic}[1]
    \STATE \COMMENT{\textnormal{\textit{Heater should be on if any room's temp is low.}}}
    \STATE heat\_on:-room(1,lo); room(2,lo); room(3,lo).
    \STATE \COMMENT{\textnormal{\textit{Room 1's temp with Pr(t1 = l) = 0.5}}}
    \STATE 0.5::room(1,lo).
    \STATE room(1,hi):-not room(1,lo).
    \STATE \COMMENT{\textnormal{\textit{Similar definitions of temps of rooms 2 and 3}}}
    \end{algorithmic}
    \end{minipage}
}
\end{center}    
\end{minipage}
\caption{ProbLog program fragment for the sensor placement problem.}
\label{fig:intro-example}
\end{figure}

Entropy of a distribution reflects the uncertainty in our knowledge of the actual state of a probabilistic system.  Given a set of random variables $S$, and any set of realizations $\mathbf{s}$ of the elements in $S$, the expected entropy of $S$ is defined as:
   \begin{quote}
       $H(S) = - \sum_{\mathbf{s}}\textit{Pr}(\mathbf{s})\log_{2}\textit{Pr}(\mathbf{s})$
   \end{quote} 
The expected entropy over the values of \texttt{heat\_on} is 
\textbf{$0.8$}.  

We can make a better estimate of whether to turn the heat on or not if we measure one or more rooms' temperature(s).  Among all random variables in a program, we may be able to obtain the valuation (i.e. \emph{measure}) of only a subset.  We mark such variables as \emph{observable}.  The act of observing may incur a cost that is independent of the outcome.  We annotate observable variables and their respective costs for observation by adding special \texttt{observable/2} facts to the program.  For instance, in the sensor placement problem, if each room permits placement of a sensor, and observing each sensor consumes one unit of energy, then the added facts are:
\begin{quote}
    \ttfamily
    observable(room(1, \_), 1).
    
    observable(room(2, \_), 1).
    
    observable(room(3, \_), 1).
\end{quote}

Now, consider the case where the building's green certification allows only enough energy for \emph{two} observations (\emph{i.e.}, total cost of two energy units). 
 The expected entropy over the values of \texttt{heat\_on} under the various choices of sensors are as follows:
   \begin{quote}
       $\textit{H}(\mathtt{heat\_on}|T_{1}, T_{2}) = \textit{H}(\mathtt{heat\_on}, T_{1}, T_{2}) - \textit{H}(T_{1}, T_{2}) = 0.31$
       
       $\textit{H}(\mathtt{heat\_on}|T_{1}, T_{3}) = \textit{H}(\mathtt{heat\_on}, T_{1}, T_{3}) - \textit{H}(T_{1}, T_{3}) = 0.18$
       
       $\textit{H}(\mathtt{heat\_on}|T_{2}, T_{3}) = \textit{H}(\mathtt{heat\_on}, T_{2}, T_{3}) - \textit{H}(T_{2}, T_{3}) = 0.31$
   \end{quote}
So, we should choose to measure the temperatures in rooms 1 and 3. Measuring temperatures of specific rooms by placing sensors in them results in additional information about the temperature situation inside the building, reflected by the expected entropy of $0.8$ in the uninformed case reducing in all three cases. In this example, the reduction in expected entropy is the \textbf{VoI} of the corresponding set of random variables. We choose to place sensors in rooms 1 and 3 since the VoI of the set $\{T_{1}, T_{3}\}$ is the highest at $0.62$. \hfill$\Box$
\end{example}

\noindent
\subparagraph{\textit{\underline{The Value of Information.}}}
In general, VoI is defined in terms of a \emph{utility} function.  The VoI of a set of variables is defined as the increase in the expected utility of the system due to observation of these variables. However, the observations themselves are expensive.  A central problem in such settings is to determine the variables to observe such that the increase in the expected utility of the system is maximized~\cite{howard:da,mookerjee:kde}. In temperature monitoring, we may have to schedule a sensor in a way that simultaneously minimizes uncertainty in temperature and energy expenditure \cite{deshpande:vldb}; in medical expert systems, we have to choose among several expensive diagnostic tests to become most certain about a patient's health while minimizing costs \cite{turney:jair};  in active learning of partially hidden Markov models, we have to label a subset of the hidden states that simultaneously optimizes predictive power of the model and the cost of annotation \cite{scheffer:ecml}. This general setting of choosing observations in an uncertain system was first introduced in \cite{howard:ssc}.

Krause and Guestrin generalized the notion of utilities to reward functions, and studied the problem of optimizing VoI in probabilistic graphical models~\cite{krause:ijcai}.  They showed that the problem is wildly intractable in general, and gave polynomial time solutions for chain graphical models.  They considered two settings of \emph{subset selection}, where we select the variables to observe \emph{a priori}; and \emph{conditional planning}, where we interleave variable selection and observation, with the variables chosen later conditioned on the outcomes of variables observed earlier.  A more detailed description of related work appears in Section~\ref{sec:discuss}.

\noindent
\subparagraph{\textit{\underline{A Greedy Approach to VoI Optimization.}}}
Given a ProbLog program, a set of \texttt{observable/2} facts, and a total \emph{budget}, we proceed in a greedy manner.  At each step, we select a single observation that has the highest \emph{VoI}.  We build a solution by recursively solving the optimization problem, removing the selected observation from further consideration, and under a suitably reduced budget, until we exhaust observations or budget.  When the subproblems are conditioned on the \emph{outcomes} of earlier observations, we get a decision tree, or a \emph{conditional observation plan}.

\paragraph{\textbf{Contributions.}}
The main contributions of this paper are two fold.  First, by considering systems specified as PLPs,  our setting of optimally selecting observations based on VoI is more general than that considered by any prior work.  The earlier works were limited to influence diagrams or probabilistic graphical models, both of which have static and explicit specification of conditional dependencies of variables.  Secondly,  we show that when the optimization problem is considered without a limiting budget, our algorithm can be used as an \emph{anytime algorithm}.  That is, preempting the algorithm anytime before completion provides a usable result, the results improve over time, and converge to the optimal solution.

The rest of the paper is organized as follows.  We begin with a brief recap of ProbLog and its semantics in Section~\ref{sec:prism:overview}.  In Section~\ref{sec:prism-voi} we describe the basis for VoI optimization in the setting of PLPs.  The greedy algorithm for selecting observations based on VoI is described in Section~\ref{sec:greedy-algorithm}.  We discuss related work in Section~\ref{sec:discuss} and conclude with a discussion of future work in Section~\ref{sec:future}.

\section{ProbLog and Distribution Semantics} \label{sec:prism:overview}

\paragraph{\textbf{Notational Conventions.}}
The technical development in this paper assumes familiarity with traditional logic programming concepts and terminology, including variables, terms, substitutions, ground and non-ground terms, atomic formulas, clauses, Herbrand models, etc.   See~\cite{NM:Prolog}.  We use ``clauses'' interchangeably with rules and facts.  We also use familiar notation, such as using symbols $\theta$, $\sigma$ for denoting substitutions, and using $p/n$ notation to denote an $n$-ary predicate $p$.
\hfill $\Box$

\bigskip

We present a brief, high-level overview of ProbLog \cite{deraedt:ijcai,kimmig:iclp}, a probabilistic extension of Prolog. A ProbLog theory $\mathcal{T}$ consists of a set of labeled facts $\mathcal{F}$, and a set of definite clauses $\mathcal{BK}$ that represent the background knowledge. The facts $p_{i}::F_{i}$ in $\mathcal{F}$ are annotated with a probability $p_{i}$ which states that $F_{i}\theta$ is true with probability $p_{i}$ for all substitutions $\theta$ grounding $F_{i}$. These random variables are assumed to be mutually independent. A ProbLog theory describes a probability distribution over the set of Prolog programs $\mathcal{L} =\mathcal{F}_{\mathcal{L}} \cup \mathcal{BK}$, where $\mathcal{F}_{\mathcal{L}} \subseteq \mathcal{F}\Theta$ and $\mathcal{F}\Theta$ denotes the set of all possible ground instances of facts in $\mathcal{F}$. If $f_{i}$ denotes  a possible grounding of any $F_{i}$ in $\mathcal{F}$, then
\begin{quote}
    $\textit{Pr}(\mathcal{L}|\mathcal{T}) = \prod_{(f_{i} \in \mathcal{F}_{\mathcal{L}})}p_{i}\prod_{(f_{i} \in \mathcal{F}\Theta\backslash\mathcal{F}_{\mathcal{L}})}(1 - p_{i})$
\end{quote}
The success probability of a query \texttt{q} is then
\begin{quote}
    $\textit{Pr}(\texttt{q}|\mathcal{T}) = \sum_{\mathcal{L}}\textit{Pr}(\texttt{q}|\mathcal{L})\textit{Pr}(\mathcal{L}|\mathcal{T})$
\end{quote}
Here $\textit{Pr}(\texttt{q}|\mathcal{L})$ is $1$ if there is a substitution $\theta$ such that $\mathcal{L}$ entails $\texttt{q}\theta$, and $0$ otherwise. Note that each $\mathcal{F}_{\mathcal{L}} \subseteq \mathcal{F}\Theta$ can be extended into a possible world by computing the least Herbrand model of $\mathcal{L} = \mathcal{F}_{\mathcal{L}} \cup \mathcal{BK}$. Thus, $\mathcal{T}$ defines a probability distribution over a set of least Herbrand models $\mathcal{M}$. From this perspective, the probability of the query \texttt{q} succeeding is the sum of the probabilities of the least models in $\mathcal{M}$ where at least one grounding $\texttt{q}\theta$ is included. This design choice was first introduced as the \emph{distribution semantics} in another probabilistic extension of Prolog, PRISM \cite{sato:ijcai,sato:jair}.

\begin{figure}
\centering
\begin{minipage}{0.6\columnwidth}
\begin{center}
\framebox{
    \begin{minipage}{\columnwidth}
    \ttfamily\small
    \begin{algorithmic}[1]
        \STATE \COMMENT{$\mathcal{F}$:}
        \STATE 0.1::tb\_prior(X):- person(X).
        \STATE 0.4::tr(X,Y):- person(X), person(Y).
        \STATE 0.3::x\_ray(X,0).
        \STATE 0.9::x\_ray(X,1).
        \STATE \COMMENT{$\mathcal{BK}$:}
        \STATE tb(X,1):- tb\_prior(X).
        \STATE tb(X,1):- friend(X,Y), tr(Y,X), tb(Y,1).
        \STATE tb(X,0):- not(tb(X,1)).
        \STATE diagnosis(X):- tb(X,D), x\_ray(X,D).
        \STATE epidemic:- findall(X,tb(X,1),E),
        \STATE \hspace{52pt}length(E,N), N>2.
        \STATE person(1).
        \STATE person(2).
        \STATE person(3).
        \STATE person(4).
        \STATE friend(1, 2).
        \STATE friend(2, 1).
        \STATE friend(2, 3).
        \STATE friend(3, 2).
        \STATE friend(3, 4).
        \STATE friend(4, 3).
    \end{algorithmic}
    \end{minipage}
}
\end{center}    
\end{minipage}
\caption{ProbLog program modeling the spread of tuberculosis among a population.}
\label{fig:prism-diagnosis}
\end{figure}

\begin{example}[\textnormal{\textbf{Tuberculosis Outbreak}}]\label{ex:epi} \rm
   The program in Figure \ref{fig:prism-diagnosis} models how tuberculosis spreads in a population of $4$. A person can develop tuberculosis on their own (line 7), with likelihood $0.1$ (line 2). A person can also develop tuberculosis if they are friends with someone who has the condition (line 8), with likelihood $0.4$ (line 3). It is possible to diagnose tuberculosis by performing an x-ray of the chest (line 10): the likelihoods of a positive diagnosis in the absence and presence of tuberculosis are respectively $0.3$ (line 4) and $0.9$ (line 5). Note that although the diagnosis is a \emph{dependent} random process, dependent on the presence or absence of tuberculosis, it is modeled as two \emph{independent} random processes, \texttt{x\_ray(0)} and \texttt{x\_ray(1)}, to maintain consistency with the assumptions of ProbLog. There is an epidemic of tuberculosis if at least $3$ persons have the condition (lines 11-12).\hfill$\Box$
\end{example}

\section{VoI Under Distribution Semantics} \label{sec:prism-voi}

We start off with an example of how VoI fits into the PLP setting.

\begin{example}[\textnormal{\textbf{Is there an Epidemic?}}] \label{ex:dec}\rm
   In Example \ref{ex:epi}, we need to become more sure about whether, or not a tuberculosis epidemic has broken out. We can perform a chest x-ray on each person to gain more information about the epidemic situation. From a VoI perspective, the utility of the system is the expected entropy over the truth values of \texttt{epidemic}, and the outcomes of the x-rays are observations. There is a deadline for declaring an epidemic that allows us only enough time to track down and perform an x-ray on only one person. So, we track down person \texttt{2} and perform the x-ray on them since it gives us the highest VoI of $0.08$: the initial expected entropy of $0.45$ comes down to $0.37$.\hfill$\Box$
\end{example}

As in Example \ref{ex:dec}, it generally makes sense to define VoI in the PLP setting in terms of a PLP and a query (\texttt{epidemic} in Example \ref{ex:dec}): we gather information to answer the query most effectively. So, it also makes sense to define utility as some function of the probability distribution over the possible groundings, or truth values of the query (expected entropy in Example \ref{ex:dec}). However, the distribution semantics of ProbLog is defined with respect to the associated least Herbrand models, not the groundings of a query. Hence, a probability distribution may not even exist for the groundings of a query. For example, given the query \texttt{q(X)}, there are $2$ least models where $X$ is simultaneously $0$ and $1$ in the following ProbLog theory:
\begin{quote}
    \ttfamily\small
    0.5::q(0).
    
    0.5::q(1).
    
    q(1):-not(q(0)).
    
    q(0):-not(q(1)).
\end{quote} 
There are, of course, ProbLog theories and queries where probability distributions exist over groundings. However, in order to make the following discussion more general, we only consider ground queries in the rest of the paper.

Before formally defining utility and VoI in ProbLog theories, we define some key concepts in Definitions \ref{def:observable} through \ref{def:scenario}.

\begin{definition}[\textnormal{\textbf{Observable}}] \label{def:observable}
Let $\mathcal{T}$ be a ProbLog theory, $\mathcal{M}$ be the set of associated least Herbrand models, and \texttt{t} be an atomic formula in $\mathcal{T}$. Let $\texttt{t}\theta$ be an arbitrary instance of \texttt{t}. Let $\sigma_{i}$ denote a substitution for $\texttt{t}\theta$ such that $\texttt{t}\theta\sigma_i$ is ground.  Let  $\mathcal{M}_{i}$ be the subset of $\mathcal{M}$ where all least models include $\texttt{t}\theta\sigma_{i}$. Then $\texttt{t}\theta$ is an observable in $\mathcal{T}$ if
\begin{itemize}[leftmargin=20pt]
    \item $\not\exists\sigma_{i}$ such that $\mathcal{M}_{i} {=} \mathcal{M}$,
    
    \item $\exists\sigma_{i}$ such that $\mathcal{M}_{i} {\neq} \emptyset$, and
    
    \item whenever $\mathcal{M}_{i_{1}} {\cap} \mathcal{M}_{i_{2}} \neq \emptyset$, $\sigma_{i_{1}} = \sigma_{i_{2}}$.   
\end{itemize}
\end{definition}

\emph{Observables} are the set of atomic formulae whose groundings can be used to distinguish between models in a program's semantics.  For instance, in Example~\ref{ex:voi}, \texttt{heat\_on}, as well as \texttt{room(1, X)}, \texttt{room(2, X)}, etc. are all observables.   Their ground instances hold in some models and not others.
Multiple instances of the same atomic formula can be considered observables in the same theory $\mathcal{T}$. For instance, in the same example, \texttt{room(1,X)} as well as \texttt{room(1,hi)} are both observable.    It is, however, not possible for a fact in $\mathcal{T}$ to be an observable, like \texttt{person(1)} in Figure \ref{fig:prism-diagnosis}, since it will hold in every least model.  It is also not possible for an instance of an atomic formula in $\mathcal{T}$, that is not a part of any least model in $\mathcal{M}$, to be considered an observable, like \texttt{tb(5, D)}, or \texttt{diagnosis(5)} in Figure \ref{fig:prism-diagnosis}. 

\begin{definition}[\textnormal{\textbf{Realization}}] \label{def:realization}
Let $\mathcal{T}$ be a ProbLog theory, $\mathcal{M}$ be the set of associated least Herbrand models, and \texttt{t} be an atomic formula in $\mathcal{T}$. Let $t\theta$ be an instance of \texttt{t} which is an observable in $\mathcal{T}$. If $\texttt{t}\theta$ is ground, either $\texttt{t}\theta$, or $\neg\texttt{t}\theta$ is a realization for $\texttt{t}\theta$. Otherwise, if $t\theta\sigma_{i}$ is ground, and $\mathcal{M}_{i}$ is the subset of $\mathcal{M}$ where all least models include $\texttt{t}\theta\sigma_{i}$, $\texttt{t}\theta\sigma_{i}$ is a realization for $\texttt{t}\theta$ if $\mathcal{M}_{i} \neq \emptyset$.
\end{definition}

When an observable is viewed as a random variable, its realization is the corresponding valuation.
The restriction on what ground instance of $t\theta$ in the above definition can be considered as its realization is reasonable: if there are no least models in $\mathcal{M}$ that includes a given ground instance of $t\theta$, then that instance will never manifest in reality.

\begin{definition}[\textnormal{\textbf{Observation}}] \label{def:observation}
Let $\mathcal{T}$ be a ProbLog theory, $\mathcal{M}$ be the set of associated least models, and \texttt{t} be an atomic formula in $\mathcal{T}$. Let $t\theta$ be an arbitrary instance of \texttt{t}, and $\texttt{t}\theta$ be an observable in $\mathcal{T}$. Then observation of $\texttt{t}\theta$ is an operation that creates a new theory $\mathcal{T}'$ by adding a realization of $\texttt{t}\theta$ to $\mathcal{T}$ as evidence.
\end{definition}

\sloppypar
If an observable is ground, its observation leads to the creation of a new theory based on its truth value. For example in Figure \ref{fig:prism-diagnosis}, observation of \texttt{diagnosis(2)} leads to $\mathcal{T}'$ being either $\mathcal{T} \cup \texttt{\{evidence(diagnosis(2), true)\}}$, or $\mathcal{T} \cup \texttt{\{evidence(diagnosis(2), false)\}}$. Again, if the observable is not ground, if there are $g$ possible realizations, its observation will lead to the creation of one among $g$ possible new theories. For example, in Figure \ref{fig:prism-diagnosis}, observation of \texttt{tb(1, D)} leads to $\mathcal{T}'$ being either $\mathcal{T} \cup \texttt{\{evidence(tb(1, 0), true)\}}$, or $\mathcal{T} \cup \texttt{\{evidence(tb(1,$ $1), true)\}}$. The new theory will be associated with all the least models in $\mathcal{M}$ that include the evidence.

Observations consume resources: in Example \ref{ex:voi}, the resource was energy, in Example \ref{ex:dec}, the resource was time. So, the definition of an observable is incomplete without the detail of the amount of resources consumed by selecting it for observation. Also, there can be observables which are not actually observable in reality. For example, in Example \ref{ex:epi}, neither the actual health condition of a person (\texttt{tb/2}), nor the presence of an epidemic (\texttt{epidemic/0}) can be directly known. We can only know about them through performing x-rays (\texttt{diagnosis/1}). So, there must be some mention in the ProbLog theory about what observables \emph{can} be selected for observation. We propose that the user specify what observables can actually be selected for observation, and the associated cost of observation, as \texttt{observable/2} facts defined as follows:

\begin{definition}[\textnormal{\texttt{observable/2}}] \label{def:obs-predicate}
Let $\mathcal{T}$ be a ProbLog theory, and \texttt{p} be a $k$-ary predicate symbol in $\mathcal{T}$. Let $\texttt{t}\theta$ be an observable in $\mathcal{T}$ with $p/k$ as its root symbol. Then to specify that $\texttt{t}\theta$ can be considered for observation, the following fact is a part of $\mathcal{T}$
\begin{quote}
    \texttt{observable($p$($\alpha_1$, $\alpha_2$, $\ldots$, $\alpha_k$), $\gamma$)}
\end{quote}
where, 
\begin{itemize}[leftmargin=20pt]
    \item each $\alpha_i$ is either a ground term or ``\texttt{\textunderscore}'' (an anonymous variable),
    
    \item p($\alpha_1$, $\alpha_2$, $\ldots$, $\alpha_k$), is a renaming of $\texttt{t}\theta$, and
    
    \item $\gamma$ is a positive real number representing the cost of observing $\texttt{t}\theta$.   
\end{itemize} 

\comment{such that
p($\alpha_1$, $\alpha_2$, $\ldots$, $\alpha_k$)
Every variable $\texttt{V}_{i}$ in \texttt{t} is replaced by either a grounding $\texttt{v}_{i}$, or `\texttt{\_}': \texttt{v$_{i}$} indicates that the corresponding variable is ground by $\theta$, while `\texttt{\_}' indicates otherwise. The quantity \texttt{cost$_{\texttt{t}}$} specifies the cost of observation for $\texttt{t}_{y}$.}
\end{definition}

In Example~\ref{ex:dec}, to specify that we can only observe the results of x-rays, and that each x-ray costs \texttt{1}, we add the following to the program in Figure \ref{fig:prism-diagnosis}:
\begin{quote}
\texttt{observable(diagnosis(1), 1)}.

\texttt{observable(diagnosis(2), 1)}.

\texttt{observable(diagnosis(3), 1)}.

\texttt{observable(diagnosis(4), 1)}.    
\end{quote}

An important point to note is that cost in Definition \ref{def:obs-predicate} is consistent with previous works on VoI where costs are associated with individual variables in the system \cite{krause:ijcai,ghosh:ictai}. However, in our PLP setting, observation of one observable does not limit the evidence gathered to that observable. For example, in Figure \ref{fig:prism-diagnosis}, on observation of \texttt{tb(2, 1)}, we know what the observation for \texttt{tb(2, \_)} will be. So, in order to make sense of costs, the ProbLog theory should be modeled such that observations are independent. For example, in Figure \ref{fig:prism-diagnosis}, adding both \texttt{observable(tb(2, \_), cost)} and \texttt{observable(tb(2, 1),~cost)} to the program would be inconsistent with reality, \emph{and} a bad design choice. It should be noted that we intend the list of \texttt{observable/2} facts to remain consistent across the various theories created through observations. We use the list as a look-up table to determine whether an observable in any particular theory can be selected for observation.

In this paper, when optimizing the selection of observables based on VoI, we sequentially select the observables, letting our choice depend on previous observations \cite{krause:jair}. This approach is different from that described in Example \ref{ex:voi}, where a subset of observables is chosen in one go. The former setting is referred to as \emph{conditional planning}, and the latter as \emph{subset selection}
~\cite{krause:ijcai}. Conditional planning can be thought of as a sophisticated version of subset selection. So, in a conditional plan, we consider an observable in the context of all the observations that have been made thus far.  The effects of observations are captured by \emph{scenarios}, defined below.

\begin{definition}[\textnormal{\textbf{Scenario}}] \label{def:scenario}
Let $\mathcal{T}$ be a ProbLog theory, and $\mathcal{A}$ be the set of all observables in $\mathcal{T}$ that are specified as \texttt{observable/2} facts. Let subset $\mathcal{O}$  of $\mathcal{A}$ be chosen for observation, and $o$ denote a set of realizations for the elements in $\mathcal{O}$. Then $\mathcal{T}$ is said to be the scenario $\mathcal{S}_{\{\}}$, and $\mathcal{T} \cup o$ is said to be the scenario $\mathcal{S}_{o}$.
\end{definition}

Thus, scenarios are \emph{created by observations}. The specific scenario created depends on the reality in which the observations are made, where realities are defined as follows.

\begin{definition}[\textnormal{\textbf{Reality}}] \label{def:world}
Let $\mathcal{T}$ be a ProbLog theory, $\mathcal{A}$ be the set of all observables in $\mathcal{T}$ that are specified as \texttt{ob\-servable/2} facts. A reality $\mathcal{R}_{a}$ is a set of realizations $a$ for the elements in $\mathcal{A}$. If any element in $\mathcal{A}$ is chosen for observation in reality $\mathcal{R}_{a}$, its realization will be as in $a$.
\end{definition}

So, any reality may be a part of multiple least models, but any least model includes only one reality, ensuring that a probability distribution exists over realities as well. 

In ProbLog, we define utility for sets of observables for a query \texttt{q} in a baseline scenario $\mathcal{S}_{b}$. Since we will not know the realizations before actually making the observations, expectation is taken over all possible sets of realizations. The formal definition is as follows.

\begin{definition}[\textnormal{\textbf{Value of Information of Observables}}] \label{def:voi-obs}
Let $\mathcal{T}$ be a ProbLog theory, \texttt{q} be a query, $\mathcal{A}$ be all observables in $\mathcal{T}$ specified as \texttt{observa\-ble/2} facts, $\mathcal{O}$ be any subset of $\mathcal{A}$, and $o$ denote any set of realizations for elements in $\mathcal{O}$. The VoI of $\mathcal{O}$ with respect to \texttt{q} in a scenario $\mathcal{S}_{b}$ is then defined as
\begin{quote}
    $\textit{VoI}(\mathcal{O}, \texttt{q}, \mathcal{S}_{b}) = \sum_{o}\textit{Pr}(o)\textit{Utility}(\texttt{q}, \mathcal{S}_{b \cup o}) - \textit{Utility}(\texttt{q}, \mathcal{S}_{b})$,
\end{quote}
where $\textit{Utility}(\texttt{q}, \mathcal{S}_{s})$ is the utility function defined on the probability distribution over the truth values of \texttt{q} in scenario $\mathcal{S}_{s}$.
\end{definition}

Based on the intended use of the ProbLog theory, the \textit{Utility} function can be defined in a number of different ways. Interesting notions of the \textit{Utility} function can be found in~\cite{krause:jair}. We present here two possible definitions for \textit{Utility}.
\begin{enumerate}[label=(\alph*)]
    \item \textbf{Reduce Uncertainty:} This definition is useful when the goal is to reduce the uncertainty around whether, or not the query \texttt{q} is \texttt{true}, and is specified as
    \begin{quote}
        $\textit{Utility}(\texttt{q}, \mathcal{S}_{s}) = \textit{Pr}(\texttt{q}|s)\log\textit{Pr}(\texttt{q}|s) + \textit{Pr}(\texttt{not(q)}|s)\log\textit{Pr}(\texttt{not(q)}|s)$.
    \end{quote}
    The utility here is the negation of the entropy associated with the distribution over the truth values of \texttt{q}. So, VoI of $\mathcal{O}$ in this case will be the \emph{reduction in expected entropy}. We use this definition in Example \ref{ex:dec}.
    
    \item \textbf{Choose Better Action:} This definition is useful when we have to choose from a set of actions, $\alpha$, whose utilities depend on the truth value of the query \texttt{q}. Given a function $U$ that defines a mapping from an action in $\alpha$, and a truth value of \texttt{q}, to a real number, it is specified as
    \begin{quote}
        $\textit{Utility}(\texttt{q}, \mathcal{S}_{s}) = \mathop{max}_{a \in \alpha}[\textit{Pr}(\texttt{q}|s)U(a, \texttt{q}) + \textit{Pr}(\texttt{not(q)}|s)U(a, \texttt{not(q)})]$.
    \end{quote}
    This definition is related to utility nodes in influence diagrams \cite{howard:da}, and the principle of maximum expected utility from decision theory \cite{howard:ssc}. VoI of $\mathcal{O}$, in this case, will be the \emph{increase in maximum expected utility}.
\end{enumerate}

In addition to VoI for observables, we can also define VoI of entire conditional observation plans as follows.

\begin{definition}[\textnormal{\textbf{Value of Information of Observation Plans}}]\label{def:voi-plan}
Let $\mathcal{T}$ be a ProbLog theory, \texttt{q} be a query, $\mathcal{A}$ be all observables in $\mathcal{T}$ specified as \texttt{observa\-ble/2} facts, and $a$ denote any set of realizations for elements in $\mathcal{A}$. Let $\pi$ denote any observation plan where elements from $\mathcal{A}$ are chosen sequentially, based on previous observations. Let $\pi(\mathcal{R}_{a})$ denote the set of realizations from observations made by plan $\pi$ in reality $\mathcal{R}_{a}$. The VoI of the plan $\pi$ with respect to \texttt{q} is then defined as
\begin{quote}
    $\textit{VoI}(\pi, \texttt{q}) = \sum_{\mathcal{R}_{a}}\textit{Pr}(\mathcal{R}_{a})\textit{Utility}(\texttt{q}, \mathcal{S}_{\pi(\mathcal{R}_{a})}) - \textit{Utility}(\texttt{q}, \mathcal{S}_{\{\}})$.
\end{quote}
\end{definition}

We describe algorithms for optimizing choice of observables based on VoI in a PLP setting in the next section.

\section{Greedy Optimization of VoI} \label{sec:greedy-algorithm}

Based on Definition \ref{def:voi-plan}, given a limit $B$ on the sum of the costs of observations (also called a \emph{budget}), we would ideally want an optimal conditional observation plans $\pi^{*}$ where
\begin{quote}
    $\pi^{*} = \mathop{argmax}_{\pi \in \Pi_{B}}\textit{VoI}(\pi, \texttt{q})$
\end{quote}
Here, $\Pi_{B}$ denotes the set of all those conditional observation plans where, in no reality, the sum of costs of observations made exceed the budget $B$. But optimizing VoI in any uncertain system has been shown to be extremely hard \cite{krause:jair,chen:jair}. Although non-myopic (non-greedy) optimization algorithms have been proposed for restricted classes of probabilistic graphical models \cite{krause:ijcai,krause:jair,ghosh:ictai}, majority of optimization approaches are myopic (greedy) \cite{vdgaag:aisbq,dittmer:uai}. Since the ProbLog programs we consider are not restricted to modeling particular types of graphical models, we propose an algorithm that, instead of $\pi^{*}$, creates a plan $\pi'$ by greedily choosing each of its steps based on Definition \ref{def:voi-obs}.

Figure \ref{fig:greedy-heuristic} shows the outline of the algorithm, which takes the initial scenario $\mathcal{S}_{\{\}}$, the query \texttt{q}, and the budget $B$ as inputs, and offers the observation plan $\pi'$ as the output. The plan $\pi'$ is computed as a greedy decision tree where each node has the following attributes:
\begin{itemize}[label=$-$]
    \item \textbf{scenario:} The scenario associated with the node.
    \item \textbf{budget:} The resources available for observations in \textbf{scenario}.
    \item \textbf{choice:} The observable to be chosen in \textbf{scenario}.
    \item \textbf{next:} The list of children of the node.
\end{itemize}

\begin{figure}
\centering
\begin{minipage}{0.55\columnwidth}
\begin{center}
\framebox{
    \begin{minipage}{\columnwidth}
    \begin{algorithmic}[1]
        \REQUIRE initial scenario: $\mathcal{S}_{\{\}}$, query: \texttt{q}, budget: $B$
        \ENSURE observation plan: $\pi'$
        \STATE $\textit{root\_node}, \textit{WL} \leftarrow \textit{\textbf{dt\_node}}(\mathcal{S}_{\{\}}, B), \textit{queue}()$
        \STATE $\textit{\textbf{enqueue}}(\textit{WL}, \textit{root\_node})$
        \WHILE{$\textit{WL} \neq \emptyset$}
            \STATE $\textit{node} \leftarrow \textit{\textbf{dequeued}}(\textit{WL})$
            \STATE $\mathcal{S}_{n}, B_{n} \leftarrow \textbf{scenario}[\textit{node}], \textbf{budget}[\textit{node}]$
            \STATE $\mathcal{A}_{n} \leftarrow \textnormal{all observables in } \mathcal{S}_{n}$
            \STATE $\mathcal{C}_{n} \leftarrow \mathop{argmax}_{\mathcal{C} \in \mathcal{A}_{n}, \mathop{cost}(\mathcal{C}) \leq B_{n}}\textit{VoI}(\{\mathcal{C}\}, \texttt{q}, \mathcal{S}_{n})$
            \IF{$\mathcal{C}_{n} \neq \textnormal{NIL}$ \textbf{and} $\textit{VoI}(\{\mathcal{C}_{n}\}, \texttt{q}, \mathcal{S}_{n}) > 0$}
                \STATE $\mathbf{choice}[\textit{node}] \leftarrow \mathcal{C}_{n}$
                \FOR{\textbf{each} realization $c_{n}$ \textbf{of} $\mathcal{C}_{n}$}
                    \STATE $\textit{new\_node} \leftarrow \textit{\textbf{dt\_node}}(\mathcal{S}_{n \cup \{c_{n}\}}, B_{n} - \textit{cost}(\mathcal{C}_{n}))$
                    \STATE $\mathbf{next}[\textit{node}][c_{n}] \leftarrow \textit{new\_node}$
                    \STATE $\textit{\textbf{enqueue}}(\textit{WL}, \textit{new\_node})$
            \ENDFOR
            \ENDIF
        \ENDWHILE
        \RETURN $\textit{root\_node}$
    \end{algorithmic}
    \end{minipage}
}
\end{center}    
\end{minipage}
\caption{Greedy algorithm for optimizing VoI in a ProbLog program.}
\label{fig:greedy-heuristic}
\end{figure}

A decision tree node is created by the function \textit{\textbf{dt\_node}}. We initially start off with the tree having just the root node. The scenario associated with the root node is $\mathcal{S}_{\{\}}$, and the budget associated with it is the initial budget $B$. While building the tree, we maintain a list of leaves in a queue \textit{WL}. The queue \textit{WL} initially only has the root node. In each step involved in expanding the tree, the first leaf node, \textit{node}, in \textit{WL} is dequeued (line 4). The scenario associated with \texttt{node} has the set of observations $n$ (following Definition \ref{def:scenario}), and $\mathcal{A}_{n}$ denotes all the observables in $\mathcal{S}_{n}$ that are specified by the \texttt{observable/2} facts (lines 5-6). Of all the observables in $\mathcal{A}_{n}$, $\mathcal{C}_{n}$ is chosen if its cost (denoted $\mathop{cost}(\mathcal{C}_{n})$) is less than the available budget $B_{n}$, and the its VoI is maximum (line 7). The choice $\mathcal{C}_{n}$ will, of course, be NIL if $\mathcal{A}_{n}$ is empty, or if no choices satisfy the budget constraint. If there is a choice, and the gain in utility is non-zero, the choice of observable is recorded for \texttt{node}, and its children are created and enqueued into \textit{WL} (lines 8-13). The decision tree is created in a breadth-first order, and leaves in the final decision tree satisfy \emph{at least one} of the following conditions:
\begin{enumerate}[label=(\arabic*)]
    \item \textbf{Absence of Observables:} There are no observables in the associated scenario that are specified by the \texttt{observable/2} facts.
    \item \textbf{Insufficient Budget:} There are observables available in the associated scenario that are specified by the \texttt{observable/2} facts, but the available budget is insufficient for observing any of them.
    \item \textbf{No Gain in Utility:} There are observables available in the associated scenario that are specified by the \texttt{observable/2} facts, as well as sufficient budget, but all of them have zero VoI.
\end{enumerate}

\begin{definition}[\textnormal{\textbf{Decision Tree Node Utility}}]\label{def:dtnode-utility}
Let \texttt{q} be the query, and \textit{node} be a decision tree node with $\textbf{scenario}$ $\mathcal{S}_{s}$. Then the utility of \textit{node} is defined as $\textit{Utility}(\texttt{q}, \mathcal{S}_{s})$.
\end{definition}

The following definition is derived from Definitions \ref{def:voi-plan} and \ref{def:dtnode-utility}, and helps us to understand how to interpret the decision tree from the perspective of VoI.

\begin{definition}[\textnormal{\textbf{Value of Information of Decision Trees}}]
Let $\mathcal{DT}^{(i)}$ denote the partial decision tree created by the algorithm in Figure \ref{fig:greedy-heuristic} after the $i$-th iteration, where $i \geq 0$, and $\mathcal{DT}^{(0)}$ is the partial decision tree with only the root node. The VoI of $\mathcal{DT}^{(i)}$ is then the difference between the expected utility of the leaf nodes, and the utility of the root node of $\mathcal{DT}^{(i)}$.
\end{definition}

Every decision tree $\mathcal{DT}^{(i)}$, where $i \geq 0$, corresponds to a partial plan $\pi'^{(i)}$ whose VoI is the VoI of $\mathcal{DT}^{(i)}$. The VoI of the final decision tree is the VoI of the complete observation plan $\pi'$. If the tree was created non-greedily with complete lookahead, the VoI of the final tree would be the VoI of the optimal plan $\pi^{*}$.

\begin{figure}
    \centering
    \includegraphics[width=0.5\columnwidth]{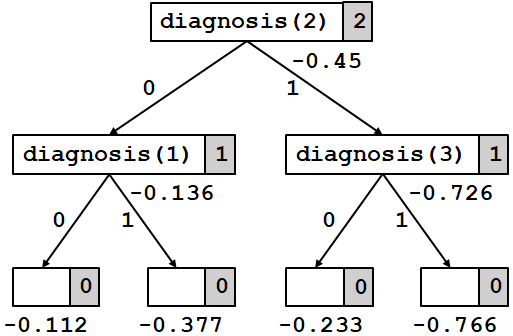}
    \caption{Decision tree created by the algorithm in Figure \ref{fig:greedy-heuristic} for the program in Figure \ref{fig:prism-diagnosis}.}
    \label{fig:decision-tree}
\end{figure}

\begin{example}[\textnormal{\textbf{Conditional Epidemic Monitoring Plan}}]\label{ex:voidec}\rm
   Figure \ref{fig:decision-tree} shows the decision tree (observation plan) created by the algorithm in Figure \ref{fig:greedy-heuristic} when the program is the one from Example \ref{ex:epi} with \texttt{observable/2} facts described in Section \ref{sec:prism-voi}, the query is \texttt{epidemic}, and the budget ($B$) is $2$. The observable chosen at a node is shown in the non-shaded part, and the budget available is shown in the shaded part. For each node, the quantity below it shows its utility defined as the negated entropy of the distribution over the truth values of \texttt{epidemic}. All the leaves in this tree satisfy only condition (2). The VoI of the final tree, and the plan it represents, is $0.081$.\hfill$\Box$
\end{example}

\paragraph{\textbf{Complexity.}}  The worst-case time complexity of the algorithm is $O(Nc_{inf}d^{N + 1})$, where $N$ is the number of observables in a program that are specified by \texttt{observable/2} facts, $d$ is the maximum number of realizations for any such observable, and $c_{inf}$ is the cost of a single probabilistic inference: the decision tree, in the worst case, has $O(d^{N})$ non-leaf nodes, and each node involves $O(Nc_{inf}d)$ worth of computation.

\paragraph{\textbf{An Anytime Algorithm.}} Following Definition \ref{def:voi-obs}, given a scenario $\mathcal{S}_{s}$, and a query \texttt{q}, $\textit{Utility}(\texttt{q}, \mathcal{S}_{s})$ does not depend on the order in which the observations in $s$ materialized. This property has an important consequence. If there is no budget ($B = \infty$), we will always end up with a plan that will be the optimal one, $\pi^{*}$, since the leaves would satisfy conditions (1) and (3), but not condition (2).  Also, the VoI of $\mathcal{DT}^{(i_{2})}$ is strictly greater than the VoI of $\mathcal{DT}^{(i_{1})}$, whenever $i_{2} > i_{1}$.   Moreover, after each extension of the tree, we get a partial plan whose VoI is strictly greater than all the previous ones.

Now, even our greedy algorithm can be considerably expensive. The preceding observations become relevant when, instead of a formal budget $B$, we have a restriction on the time available to come up with a plan. Under such a scenario, the algorithm behaves like an \emph{anytime} algorithm:  each iteration leads to a plan with strictly greater VoI, and if it runs till completion, we end up with a plan with optimal VoI.   Keeping in line with another anytime algorithm in the VoI domain \cite{horsch:uai}, we may even make \textit{WL} in Figure \ref{fig:greedy-heuristic} a priority queue with different definitions of priority: having a priority queue allows us to extend the tree in a custom order, enabling us to greedily expand parts of the plan that would lead to the greatest increase in VoI within the given time limit.

\section{Discussion}\label{sec:discuss}

VoI, as a concept, was presented by Howard~\cite{howard:ssc} in the context of decision theory. Krause and Guestrin~\cite{krause:ijcai} presented a more general notion of VoI based on reward functions defined on probability distributions of variables in the system. Also, VoI has so far been studied in either the context of influence diagrams \cite{howard:da,dittmer:uai}, or probabilistic graphical models \cite{krause:ijcai,krause:jair,ghosh:ictai}. This work presents a framework for incorporating 
the broad definition of VoI introduced in~\cite{krause:ijcai} into PLPs, which can model a wide range of probabilistic systems including both influence diagrams and probabilistic graphical models. In other words, this work presents VoI in a context that is broader than anything considered before.

Van den Broeck et al~\cite{broeck:aaai} studied optimal decision making in uncertain systems modeled in a ProbLog extension called DTProbLog. They consider actions which have utilities that depend on the least Herbrand model, and choose actions that result in the maximum expected utility. Through Definition (b) of utility in Section \ref{sec:prism-voi}, we propose how to acquire information such that the maximum expected utility is optimized. One contribution of this work can be viewed as providing an `add-on' to the functionality provided in DTProbLog. For example, in Example \ref{ex:epi}, if there is an action \texttt{declare\_epidemic}, DTProbLog would decide on whether, or not the action should be taken based on only the program given in Figure \ref{fig:prism-diagnosis}. Using our `add-on', the most informative x-rays would be performed, and the corresponding \texttt{evidence/2} (ProbLog built-in) facts would be added to the program in Figure \ref{fig:prism-diagnosis}, before the decision is made.

Planning with sensing actions and incomplete information has been studied before in a logic programming framework by Tu et al~\cite{tu:tplp}. If observing variables are thought of as sensing actions, this work seemingly also deals with a similar problem. However, they are very different. In planning under incomplete information, sensing is guided entirely by satisfying the specified goal with \emph{no restrictions} on what can be sensed, \emph{and} the resources available for sensing. In optimization of VoI, sensing is guided by acquiring the most information and there are \emph{restrictions} on what can be sensed, and \emph{possibly} the resources available for sensing. Gathering information in our framework can be viewed as adding \texttt{evidence/2} facts to a ProbLog theory. So, in a way, observations are `abducted' into the theory. However, unlike in abductive logic programming \cite{denecker:jlp}, our `abduction' results in additional information being obtained, not a goal being explained through already available information.

Quinlan first proposed optimizing VoI through myopic (greedy) decision trees~\cite{quinlan:ml}. Many subsequent works have also proposed greedy approaches to optimizing VoI~\cite{vdgaag:aisbq,dittmer:uai,krause:corr}. Krause and Guestrin showed that the VoI optimization problem even in very restricted graphical models is extremely hard, belonging to complexity classes at least as hard as $\mathbf{NP^{PP}}$-complete~\cite{krause:jair}.  
Efficient VoI optimization algorithms were described in~\cite{krause:ijcai} and~\cite{ghosh:ictai}, but their efficiency is restricted to just chain-like graphical models. So, given the hardness of the domain \emph{and} the broadness of our context, the simplistic greedy algorithm in Figure \ref{fig:greedy-heuristic} is a good starting point, and its anytime nature in the absence of budgets is meaningful. The algorithm provides a very broad VoI optimization framework. For example, instead of using the definitions of utility in Section \ref{sec:prism-voi} along with a budget, we can simply use the same-decision probability \cite{chen:jair} as the combined selection and stopping criteria.

\section{Conclusions and Future Work}\label{sec:future}

We have presented a framework for incorporating a broad notion of VoI in the general setting of PLPs. Following initial approaches to VoI in other contexts, such as influence diagrams \cite{vdgaag:aisbq} and graphical models \cite{heckerman:pami}, we have presented a greedy approach for selecting observations based on VoI in systems modeled as PLPs. The optimal algorithms of~\cite{krause:ijcai} are highly efficient for chain graphical models, but cannot be applied to any other kinds of models. The optimal algorithms of~\cite{ghosh:ictai} are applicable to general dynamic Bayesian networks, while being more efficient than the algorithms of~\cite{krause:ijcai} for chain models. PLPs provide an expressive framework where we can model much more than just probabilistic graphical models. Our eventual goal is to come up with optimal algorithms for PLPs that, like the algorithms in~\cite{ghosh:ictai}, naturally achieve efficiency on PLPs that encode chain models. We expect this will involve exposing the conditional dependence between random variables in the PLP. We believe that the explanation graphs used in inference contain adequate information for exposing these dependencies. Although relatively few, there are approximation algorithms that offer a priori \cite{krause:corr} and a posteriori \cite{ghosh:ictai} approximation bounds. We would like to explore whether such approximation algorithms are possible in the context of PLPs.

\bibliographystyle{eptcs}
\bibliography{tlp2esam}

\label{lastpage}
\end{document}